\documentclass{vllm}
\usepackage[margin=1.2in]{geometry}
\usepackage{amsmath,amssymb,amsthm}
\usepackage{mathtools}
\usepackage{tikz}
\usetikzlibrary{shapes.geometric,arrows.meta,positioning,fit,backgrounds,
                calc,decorations.pathreplacing,matrix,shadows,patterns}
\usepackage{pgfplots}
\pgfplotsset{compat=1.18}

\usepackage[utf8]{inputenc}
\usepackage[T1]{fontenc}
\usepackage{lmodern}
\usepackage{graphicx}
\usepackage{booktabs}
\usepackage{hyperref}
\usepackage{cleveref}
\usepackage{listings}
\usepackage{xcolor}
\usepackage{algorithm}
\usepackage{algpseudocode}
\usepackage{enumitem}
\usepackage{tikz}
\usetikzlibrary{shapes.geometric,arrows.meta,positioning,fit,calc,
  decorations.pathreplacing}
\usepackage{pgfplots}
\pgfplotsset{compat=1.18}
\usepackage{caption}
\usepackage{subcaption}
\usepackage{multirow}
\usepackage{makecell}
\usepackage{pifont}
\usepackage{mathtools}

\hypersetup{
  colorlinks=true,
  linkcolor=blue!70!black,
  citecolor=green!50!black,
  urlcolor=blue!60!black,
}

\definecolor{codebg}{HTML}{F5F5F5}
\definecolor{codeframe}{HTML}{CCCCCC}
\newcommand{\Lin}{L_{\text{in}}}
\newcommand{\Lout}{L_{\text{out}}}
\newcommand{\Ltotal}{L_{\text{total}}}
\newcommand{\Bshort}{B_{\text{short}}}
\newcommand{\Ps}{\mathcal{P}_s}
\newcommand{\Pl}{\mathcal{P}_l}

\title{%
  Dual-Pool Token-Budget Routing for Cost-Efficient and Reliable LLM Serving%
}

\author{
  Xunzhuo Liu$^{1}$, Bowei He$^{1,2,3,\dagger}$, Xue Liu$^{1,2,3,4}$, Andy Luo$^{5}$, \\ Haichen Zhang$^{5}$, Huamin Chen$^{1}$ \\
  $^1$ vLLM Semantic Router Project, $^2$ MBZUAI, $^3$ McGill University, \\$^4$Mila, $^5$ AMD \\
  $^\dagger$ Corresponding author: \texttt{Bowei.He@mbzuai.ac.ae}
}
\date{April 2026}

\abstract{%
Production vLLM fleets typically provision each instance for the worst-case context length, leading to substantial KV-cache over-allocation and under-utilized concurrency. In practice, 80--95\% of requests are short, yet are served under configurations optimized for long contexts, wasting 4--8$\times$ throughput capacity and triggering reliability issues such as OOM crashes, preemption, and request rejections. We identify a common root cause for these inefficiencies: \emph{configuration--traffic mismatch}. We propose \emph{dual-pool token-budget routing}, a lightweight dispatch mechanism that partitions a homogeneous fleet into two specialized pools: a high-throughput short-context pool and a high-capacity long-context pool. Each request is routed based on its estimated total token budget, computed using a per-category bytes-to-token ratio that is learned online via exponential moving average from \texttt{usage.prompt\_tokens} feedback, eliminating the need for a tokenizer. We also develop a simple analytical model that predicts fleet-level cost savings from workload characteristics and measured throughput differences, enabling practitioners to estimate benefits prior to deployment. Evaluations on real-world traces from the Azure LLM Inference Dataset and LMSYS-Chat-1M, serving Llama-3-70B on A100 GPUs, show that our approach reduces GPU-hours by \textbf{31--42\%}, corresponding to \$2.86M annual savings at fleet scale, while lowering preemption rates by \textbf{5.4$\times$} and improving P99 TTFT by 6\%. A case study with Qwen3-235B-A22B on AMD MI300X at 10{,}000~req/s projects \textbf{\$15.4M} in annual savings. The method incurs only $O(1)$ dispatch overhead, adapts automatically to heterogeneous workloads, and composes seamlessly with existing optimizations such as PagedAttention, continuous batching, and prefill--decode disaggregation.
}

\begin{document}
\maketitle

\section{The Problem: One Pool, Two Failures}
\label{sec:problem}

\subsection{Homogeneous Provisioning Wastes GPUs}

The standard vLLM deployment configures every instance for the longest
context window any request might need.
Analysis of the Azure LLM Inference Dataset~\cite{azure-llm-traces}
reveals that 80\% of requests fit in 2K tokens and 95\% fit in 8K,
yet fleets are configured for \texttt{max\_model\_len}=64K+.
The LMSYS-Chat-1M corpus~\cite{zheng2024lmsyschat1m} reports a mean
prompt length of just 69.5 tokens.
Independent production traces confirm this pattern:
BurstGPT's 10.3M-request Azure dataset shows the same
short-dominated distribution~\cite{burstgpt2024}, and
Alibaba's ServeGen characterization of billions of cloud requests
finds input lengths follow a Pareto/log-normal mixture heavily
concentrated below 2K tokens~\cite{servegen2025}.

This matters because \texttt{max\_model\_len} directly controls
concurrency. For a model with $n_l$ layers, $n_h$ KV-heads, and head
dimension $d_h$, the KV-cache per sequence is:
\begin{equation}
  M_{\text{seq}}
  = 2 \cdot n_l \cdot n_h \cdot d_h \cdot b_{\text{dtype}}
    \cdot C_{\max}.
  \label{eq:kv-per-seq}
\end{equation}
The maximum concurrent sequences per GPU is:
\begin{equation}
  N_{\text{seq}}
  = \left\lfloor
      \frac{M_{\text{gpu}} \cdot u - M_{\text{model}}}{M_{\text{seq}}}
    \right\rfloor.
  \label{eq:max-seqs}
\end{equation}
On an 80\,GB A100 serving Llama-3-70B, $C_{\max}$=64K yields
$N_{\text{seq}} \approx 16$. Reducing to $C_{\max}$=8K yields
$N_{\text{seq}} \approx 128$ --- an
\textbf{8$\times$ concurrency gain}~\cite{vllm-optimization}.
Every short request served at the lower concurrency is pure waste.

\begin{figure}[htbp]
\centering
\begin{tikzpicture}[font=\small]
\def\barW{6.2}
\def\barH{0.55}

\node[anchor=east,font=\scriptsize,align=right] at (-0.15,-0.5)
  {Homogeneous\\[-1pt]$C_{\max}{=}$64K\\[-1pt]16 slots};
\foreach \i in {0,...,15} {
  \pgfmathsetmacro{\x}{\i * \barW / 16}
  \pgfmathsetmacro{\w}{\barW / 16}
  \fill[red!12] (\x, -0.25) rectangle +(\w, -\barH);
  \draw[gray!60] (\x, -0.25) rectangle +(\w, -\barH);
}
\foreach \i/\frac in {0/0.03,1/0.02,2/0.05,3/0.01,4/0.04,
  5/0.02,6/0.08,7/0.01,8/0.03,9/0.02,10/0.06,11/0.01,
  12/0.03,13/0.50,14/0.02,15/0.01} {
  \pgfmathsetmacro{\x}{\i * \barW / 16}
  \pgfmathsetmacro{\w}{\barW / 16 * \frac}
  \fill[red!70] (\x, -0.25) rectangle +(\w, -\barH);
}
\node[font=\scriptsize,text=red!70!black,anchor=west]
  at (\barW+0.2,-0.5) {$\approx$5\% used};

\node[anchor=east,font=\scriptsize,align=right] at (-0.15,-1.6)
  {Short pool\\[-1pt]$C_{\max}{=}$8K\\[-1pt]128 slots};
\foreach \i in {0,...,31} {
  \pgfmathsetmacro{\x}{\i * \barW / 32}
  \pgfmathsetmacro{\w}{\barW / 32}
  \fill[blue!12] (\x, -1.35) rectangle +(\w, -\barH);
  \draw[gray!40] (\x, -1.35) rectangle +(\w, -\barH);
}
\foreach \i/\frac in {0/0.25,1/0.18,2/0.40,3/0.12,4/0.30,
  5/0.15,6/0.60,7/0.10,8/0.22,9/0.18,10/0.45,11/0.08,
  12/0.25,13/0.35,14/0.15,15/0.20,16/0.30,17/0.12,18/0.55,
  19/0.08,20/0.22,21/0.38,22/0.15,23/0.28,24/0.18,25/0.42,
  26/0.10,27/0.32,28/0.20,29/0.15,30/0.25,31/0.35} {
  \pgfmathsetmacro{\x}{\i * \barW / 32}
  \pgfmathsetmacro{\w}{\barW / 32 * \frac}
  \fill[blue!70] (\x, -1.35) rectangle +(\w, -\barH);
}
\node[font=\scriptsize,text=blue!70!black,anchor=west]
  at (\barW+0.2,-1.5) {$\approx$25\% used};
\node[font=\tiny,anchor=west] at (\barW+0.2,-1.75)
  {(32 of 128 shown)};

\fill[red!70] (0,-2.25) rectangle +(0.25,0.18);
\node[font=\tiny,anchor=west] at (0.22,-2.16) {Actual KV used};
\fill[red!12] (2.4,-2.25) rectangle +(0.25,0.18);
\draw[gray!60] (2.4,-2.25) rectangle +(0.25,0.18);
\node[font=\tiny,anchor=west] at (2.62,-2.16) {Allocated (wasted)};
\end{tikzpicture}
\caption{KV-cache waste under homogeneous provisioning.
  Each slot reserves $C_{\max}$ tokens regardless of actual usage.
  Short requests (80\%+ of traffic) use $<$5\% of their allocated
  space. An 8K pool uses 5$\times$ more of each slot.}
\label{fig:kv-waste}
\end{figure}
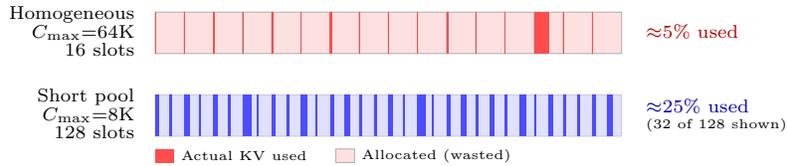

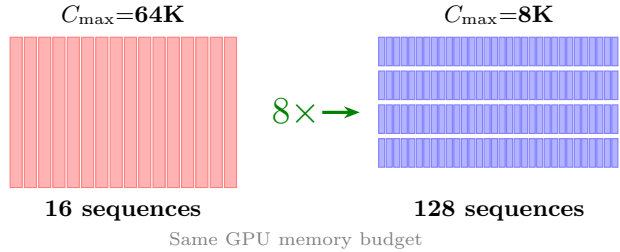
\begin{figure}[htbp]
\centering
\begin{tikzpicture}[font=\small]
\node[font=\small\bfseries] at (1.5,0.3) {$C_{\max}{=}$64K};
\foreach \i in {0,...,15} {
  \pgfmathsetmacro{\x}{\i * 0.19}
  \fill[red!30] (\x,0) rectangle +(0.16,-2.0);
  \draw[red!50] (\x,0) rectangle +(0.16,-2.0);
}
\node[font=\small\bfseries] at (1.5,-2.3) {16 sequences};

\node[font=\Large\bfseries,text=green!50!black] at (3.8,-1.0)
  {$8{\times}$};
\draw[-{Stealth[length=6pt]},very thick,green!50!black]
  (4.15,-1.0) -- (4.65,-1.0);

\node[font=\small\bfseries] at (6.5,0.3) {$C_{\max}{=}$8K};
\foreach \row in {0,...,3} {
  \pgfmathsetmacro{\ry}{-\row * 0.45}
  \foreach \i in {0,...,31} {
    \pgfmathsetmacro{\x}{4.9 + \i * 0.1}
    \fill[blue!30] (\x,\ry) rectangle +(0.08,-0.38);
    \draw[blue!50] (\x,\ry) rectangle +(0.08,-0.38);
  }
}
\node[font=\small\bfseries] at (6.5,-2.3) {128 sequences};

\node[font=\scriptsize,text=gray] at (3.8,-2.7)
  {Same GPU memory budget};
\end{tikzpicture}
\caption{Concurrency comparison: the same GPU memory holds 16
  sequences at 64K context or 128 sequences at 8K --- an
  8$\times$ difference in throughput capacity.}
\label{fig:concurrency}
\end{figure}

\begin{figure}[htbp]
\centering
\begin{tikzpicture}[font=\small]
\def\th{0.45}

\node[font=\small\bfseries,anchor=west] at (0,1.0)
  {Homogeneous pool};
\node[font=\scriptsize,anchor=east] at (-0.1,0.3) {GPU};

\fill[red!50] (0,0.1) rectangle (5.5,0.1+\th);
\node[font=\scriptsize,white] at (2.75,0.33)
  {long prefill (32K tokens)};

\fill[orange!30] (5.6,0.1) rectangle (5.95,0.1+\th);
\fill[orange!30] (6.0,0.1) rectangle (6.25,0.1+\th);
\fill[orange!30] (6.3,0.1) rectangle (6.5,0.1+\th);
\node[font=\tiny] at (6.05,0.33) {short};

\draw[decorate,decoration={brace,amplitude=4pt,mirror},gray]
  (0,-0.1) -- (5.5,-0.1)
  node[midway,below=4pt,font=\scriptsize,text=red!70!black]
  {short requests \textbf{blocked} --- TTFT violated};

\draw[-{Stealth[length=4pt]},gray] (0,-0.75) -- (7.5,-0.75)
  node[right,font=\scriptsize]{time};

\node[font=\small\bfseries,anchor=west] at (0,-1.1)
  {Token-budget routing};

\node[font=\scriptsize,anchor=east] at (-0.1,-1.6)
  {Short pool};
\fill[blue!40] (0,-1.4) rectangle (0.4,-1.4-\th);
\fill[blue!40] (0.45,-1.4) rectangle (0.8,-1.4-\th);
\fill[blue!40] (0.85,-1.4) rectangle (1.15,-1.4-\th);
\fill[blue!40] (1.2,-1.4) rectangle (1.5,-1.4-\th);
\fill[blue!40] (1.55,-1.4) rectangle (1.85,-1.4-\th);
\fill[blue!40] (1.9,-1.4) rectangle (2.2,-1.4-\th);
\node[font=\scriptsize] at (1.1,-1.62) {short requests};

\node[font=\scriptsize,anchor=east] at (-0.1,-2.35)
  {Long pool};
\fill[red!35] (0,-2.15) rectangle (5.5,-2.15-\th);
\node[font=\scriptsize,white] at (2.75,-2.37)
  {long prefill (32K tokens)};

\draw[decorate,decoration={brace,amplitude=4pt,mirror},
      green!50!black]
  (0,-2.75) -- (2.2,-2.75)
  node[midway,below=4pt,font=\scriptsize,text=green!40!black]
  {served \textbf{immediately} --- no blocking};

\end{tikzpicture}
\caption{Prefill head-of-line blocking.  \textbf{Top:} a single 32K
  prefill stalls all queued short requests.
  \textbf{Bottom:} token-budget routing isolates the long prefill in a
  separate pool; short requests proceed without delay.}
\label{fig:hol-blocking}
\end{figure}
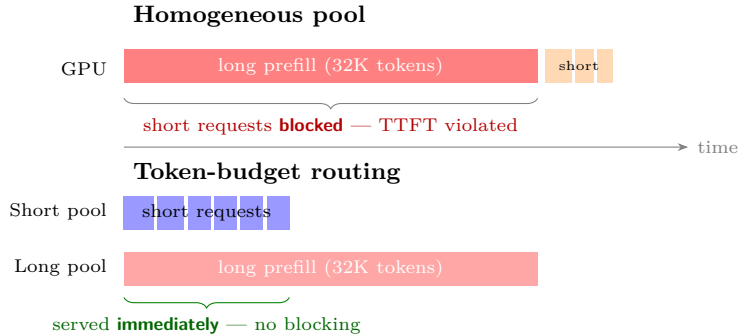

Figures~\ref{fig:kv-waste}--\ref{fig:hol-blocking} illustrate why
variable prompt lengths are particularly harmful under homogeneous
provisioning.

\subsection{Chunked Prefill: Necessary but Insufficient}
\label{sec:chunked-prefill}

vLLM's \emph{chunked prefill}~\cite{agrawal2024sarathi} mitigates
head-of-line blocking by splitting a long prefill into fixed-size
chunks (default 2{,}048 tokens) and interleaving them with decode
iterations.  This improves GPU utilization by overlapping
compute-bound prefill with memory-bound decode.

\begin{figure}[htbp]
\centering
\begin{tikzpicture}[font=\small]

\node[font=\small\bfseries,anchor=west] at (0,2.7)
  {Compute: chunked prefill interleaves with decode};

\node[font=\scriptsize,anchor=east] at (-0.1,2.1) {GPU};
\def\ch{0.5}
\fill[red!50] (0,1.9) rectangle (1.0,1.9+\ch);
\node[font=\tiny,white] at (0.5,2.15) {chunk 1};
\fill[blue!30] (1.05,1.9) rectangle (1.55,1.9+\ch);
\node[font=\tiny] at (1.3,2.15) {dec};
\fill[red!50] (1.6,1.9) rectangle (2.6,1.9+\ch);
\node[font=\tiny,white] at (2.1,2.15) {chunk 2};
\fill[blue!30] (2.65,1.9) rectangle (3.15,1.9+\ch);
\node[font=\tiny] at (2.9,2.15) {dec};
\fill[red!40] (3.2,1.9) rectangle (4.2,1.9+\ch);
\node[font=\tiny,white] at (3.7,2.15) {chunk 3};
\fill[blue!30] (4.25,1.9) rectangle (4.75,1.9+\ch);
\node[font=\tiny] at (4.5,2.15) {dec};
% ...
\node[font=\scriptsize] at (5.1,2.15) {$\cdots$};
\fill[red!30] (5.4,1.9) rectangle (6.4,1.9+\ch);
\node[font=\tiny,white] at (5.9,2.15) {chunk N};

\draw[-{Stealth[length=4pt]},gray] (0,1.75) -- (7.0,1.75)
  node[right,font=\scriptsize]{time};

\node[font=\scriptsize,text=green!50!black,anchor=west] at (0,1.45)
  {\checkmark~Decodes not blocked --- good latency};
\node[font=\small\bfseries,anchor=west] at (0,0.9)
  {Memory: KV cache still allocated for full context};
\def\mW{6.5}
\def\mH{0.6}
\fill[gray!8] (0,-0.5) rectangle (\mW,\mH-0.5);
\draw[gray!50] (0,-0.5) rectangle (\mW,\mH-0.5);
\fill[gray!35] (0,-0.5) rectangle (1.8,\mH-0.5);
\node[font=\tiny] at (0.9,-0.2) {Model weights};
\draw[gray!50] (1.8,-0.5) -- (1.8,\mH-0.5);
\fill[red!40] (1.85,-0.5) rectangle (4.4,\mH-0.5);
\node[font=\tiny,white,align=center] at (3.12,-0.2)
  {Seq 1: 32K tokens\\(still full allocation!)};
\draw[gray!50] (4.4,-0.5) -- (4.4,\mH-0.5);
\fill[blue!25] (4.45,-0.5) rectangle (4.85,\mH-0.5);
\node[font=\tiny,rotate=90] at (4.65,-0.2) {Seq 2};
\fill[blue!25] (4.9,-0.5) rectangle (5.3,\mH-0.5);
\node[font=\tiny,rotate=90] at (5.1,-0.2) {Seq 3};
\fill[green!8] (5.35,-0.5) rectangle (\mW,\mH-0.5);
\node[font=\tiny,text=gray] at (5.92,-0.2) {free};
\draw[decorate,decoration={brace,amplitude=4pt},red!70!black]
  (1.85,\mH-0.42) -- (4.4,\mH-0.42)
  node[midway,above=4pt,font=\tiny,text=red!70!black]
  {chunked compute $\neq$ chunked memory};
\node[font=\scriptsize,text=red!70!black,anchor=west] at (0,-0.85)
  {\ding{55}~Same KV footprint --- same concurrency limit --- same fleet size};
\end{tikzpicture}
\caption{Chunked prefill: compute vs.\ memory.
  \textbf{Top:} chunked prefill interleaves long-prefill chunks with
  decode iterations, preventing head-of-line blocking.
  \textbf{Bottom:} the KV cache is still allocated for the full 32K
  sequence, not the chunk size. The memory footprint, concurrency
  limit, and fleet size are unchanged.}
\label{fig:chunked-prefill}
\end{figure}
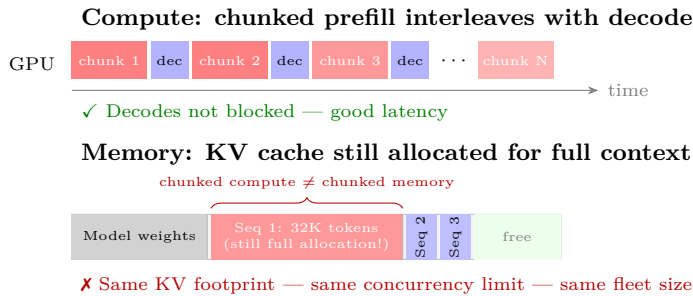

However, chunked prefill solves only the \emph{compute scheduling}
problem --- it does not address the \emph{memory provisioning}
problem (\Cref{fig:chunked-prefill}):

\begin{itemize}[nosep]
  \item \textbf{KV cache is allocated for the full sequence, not the
    chunk.}  A 32K-token request processed in 2K chunks still
    reserves 32K tokens of KV-cache capacity for the entire duration.
    The memory footprint is identical to unchunked prefill.
  \item \textbf{$C_{\max}$ still dictates concurrency.}  Every
    instance must be provisioned for the worst-case context window.
    The 8$\times$ concurrency gap between 8K and 64K configurations
    (\Cref{fig:concurrency}) remains.
  \item \textbf{Preemption and OOM persist.}  Under high load, many
    concurrent sequences with large KV footprints still exhaust the
    cache budget, triggering the same preemption storms and OOM events.
  \item \textbf{Fleet size is unchanged.}  Chunked prefill improves
    per-request latency but does not reduce the number of GPU
    instances required to serve a given throughput target.
\end{itemize}

Token-budget pool routing is \emph{complementary}: it solves the
memory problem that chunked prefill leaves open.  Each pool can
internally use chunked prefill, gaining the scheduling benefits
within a right-sized memory configuration.

\subsection{Homogeneous Provisioning Causes Failures}

The same over-provisioning that wastes money triggers reliability
failures when the fleet is pushed to high utilization:

\begin{itemize}[nosep]
  \item \textbf{OOM crashes}: bursts of medium-length requests
    collectively exceed KV-cache capacity.
  \item \textbf{Preemption storms}: vLLM evicts in-progress sequences,
    degrading throughput and tail latency~\cite{vllm-optimization}.
  \item \textbf{Request rejections}: requests exceeding
    \texttt{max\_model\_len} are dropped before inference begins.
  \item \textbf{Head-of-line blocking}: long prefills stall short
    requests, violating TTFT SLOs.
\end{itemize}

\subsection{Root Cause: Configuration--Traffic Mismatch}

These cost and reliability problems are not independent. Both stem from
a mismatch between the pool's static configuration (sized for 64K) and
the actual traffic distribution (concentrated below 8K). Eliminating
this mismatch resolves both simultaneously.

\subsection{Contributions}

We make four contributions:

\begin{enumerate}[nosep]
  \item \textbf{Token-budget pool routing} (\Cref{sec:algorithm}):
    a fleet-level dispatch algorithm that splits a homogeneous vLLM
    fleet into right-sized short and long pools with $O(1)$ overhead.
    Unlike per-GPU optimizations (PagedAttention, chunked prefill,
    speculative decoding), it operates \emph{across} instances and
    composes with all of them.
  \item \textbf{Self-calibrating token estimation}
    (\Cref{sec:estimation}): a per-category EMA that learns the
    bytes-per-token ratio from \texttt{usage.prompt\_tokens} feedback
    with asymmetric-error-aware conservative bias. This eliminates
    the need for a model-specific tokenizer at the routing layer ---
    a practical constraint in multi-model deployments where the router
    sits upstream of heterogeneous backends.
  \item \textbf{Closed-form cost model} (\Cref{sec:analysis}):
    $\text{savings} = \alpha\,(1 - 1/\rho)$, which predicts fleet-level
    GPU savings from two quantities observable \emph{before deployment}:
    the traffic CDF and profiled throughput. This lets teams audit the
    savings opportunity without changing infrastructure, in contrast to
    simulation-dependent approaches~\cite{agrawal2024vidur,sageserve2025}.
  \item \textbf{Comprehensive evaluation} (\Cref{sec:eval}):
    on two real-world traces (Azure, LMSYS) and frontier hardware
    (Qwen3-235B on MI300X), demonstrating 31--42\% GPU reduction,
    5.4$\times$ fewer preemptions, and \$15.4M/yr savings at
    10{,}000 req/s.
\end{enumerate}

\section{Token-Budget Pool Routing}
\label{sec:algorithm}

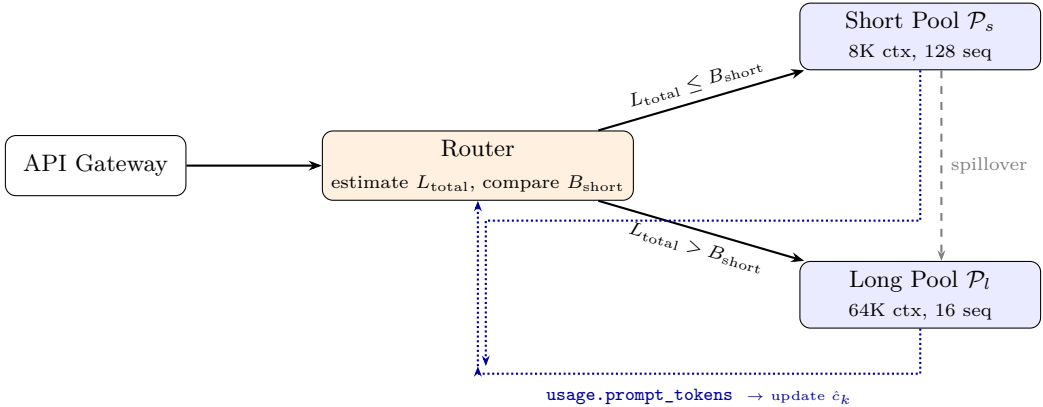
\begin{figure}[htbp]
\centering
\begin{tikzpicture}[
  node distance=0.8cm and 1.6cm,
  box/.style={draw, rounded corners=4pt, minimum width=2.4cm,
              minimum height=0.8cm, align=center, font=\small},
  pool/.style={draw, rounded corners=4pt, minimum width=3.2cm,
               minimum height=0.8cm, align=center, fill=blue!8,
               font=\small},
  arrow/.style={-{Stealth[length=5pt]}, thick},
  darrow/.style={-{Stealth[length=4pt]}, thick, dashed, gray},
  fbarrow/.style={-{Stealth[length=4pt]}, thick, densely dotted,
                  blue!60!black},
]
  \node[box] (gw) {API Gateway};
  \node[box, right=1.8cm of gw, fill=orange!12,
        minimum width=3.0cm] (rt) {%
    Router\\[1pt]
    {\scriptsize estimate $\Ltotal$, compare $\Bshort$}%
  };
  \node[pool, above right=0.8cm and 2.2cm of rt] (sp)
    {Short Pool $\Ps$\\{\scriptsize 8K ctx, 128 seq}};
  \node[pool, below right=0.8cm and 2.2cm of rt] (lp)
    {Long Pool $\Pl$\\{\scriptsize 64K ctx, 16 seq}};
  \draw[arrow] (gw) -- (rt);
  \draw[arrow] (rt) -- node[above,font=\scriptsize,sloped]
    {$\Ltotal \leq \Bshort$} (sp);
  \draw[arrow] (rt) -- node[below,font=\scriptsize,sloped]
    {$\Ltotal > \Bshort$} (lp);
  \draw[darrow] ([xshift=8pt]sp.south) -- ([xshift=8pt]lp.north)
    node[midway,right,font=\scriptsize,text=gray]{spillover};
  \coordinate (fbl) at ($(lp.south)+(0,-0.6)$);
  \coordinate (fbr) at ($(rt.south)+(0,-2.2)$);
  \draw[fbarrow] (lp.south) -- (fbl) -| (fbr);
  \draw[fbarrow] (sp.south) -- +(0,-2.0) -| ([xshift=3pt]fbr);
  \draw[fbarrow] (fbr) -- (rt.south);
  \node[font=\scriptsize, text=blue!60!black, align=center]
    at ($(fbr)!0.5!(fbl |- fbr)$) [below=5pt] {%
      \texttt{usage.prompt\_tokens}
      {\tiny ~$\to$ update $\hat{c}_k$}};
\end{tikzpicture}
\caption{Token-budget pool routing with closed-loop calibration.
  Each request's total token budget $\Ltotal$ is estimated using
  calibrated $\hat{c}_k$ and compared against threshold $\Bshort$.
  Responses feed back \texttt{usage.prompt\_tokens} to refine the
  per-category ratio. A spillover path handles burst overloads.}
\label{fig:arch}
\end{figure}

The core idea is simple: split a homogeneous fleet into two pools:
a \emph{short pool} $\Ps$ with a small \texttt{max\_model\_len}
(high concurrency, high throughput) and a \emph{long pool} $\Pl$
with the original context window (lower throughput, but handles all
requests), and route each request to the appropriate pool based
on its total token budget (\Cref{fig:arch}).

\subsection{Token-Budget Estimation}
\label{sec:estimation}

For each request $r$ with traffic category $k$ (e.g., code, prose,
CJK), the router estimates the total token budget:
\begin{equation}
  \Ltotal = \underbrace{\left\lceil |r| \;/\; \hat{c}_k \right\rceil}_{\Lin}
          + \underbrace{r.\texttt{max\_output\_tokens}}_{\Lout},
  \label{eq:budget}
\end{equation}
where $|r|$ is the request's byte length and $\hat{c}_k$ is the
\emph{calibrated} bytes-per-token ratio for category $k$.
The estimate is $O(1)$: a single division, no tokenizer required.

\paragraph{Cold start.}
Before any feedback is available, $\hat{c}_k$ defaults to $c_0{=}4.0$
(the English-prose average). This is accurate enough for routing ---
the threshold analysis in \Cref{sec:sensitivity} shows that even
moderate estimation error has little impact on savings.

\paragraph{Closed-loop calibration.}
Every LLM response includes the exact prompt token count in the
\texttt{usage.prompt\_tokens} field.  The router uses this signal to
update $\hat{c}_k$ via an exponential moving average (EMA):
\begin{equation}
  c_{\text{obs}} = \frac{|r|}{\texttt{usage.prompt\_tokens}}, \qquad
  \hat{c}_k \;\leftarrow\; \beta \,\hat{c}_k + (1{-}\beta)\,c_{\text{obs}},
  \label{eq:ema}
\end{equation}
with decay $\beta{=}0.95$.  Because routing errors are asymmetric ---
sending a long request to the short pool causes preemption, while
sending a short request to the long pool merely wastes some
throughput --- the router uses a \emph{conservative} estimate:
\begin{equation}
  \hat{c}_k^{\text{route}}
  = \hat{c}_k - \gamma\,\hat{\sigma}_k,
  \label{eq:conservative}
\end{equation}
where $\hat{\sigma}_k$ is the EMA standard deviation and
$\gamma{=}1.0$ biases toward overestimating token count
(i.e., toward the safer long pool for borderline requests).

\paragraph{Why per-category.}
A single global ratio is a poor fit for heterogeneous traffic:
code averages ${\sim}3.5$ bytes/token, CJK text ${\sim}2.0$, and
English prose ${\sim}4.5$.  Recent work confirms that tokenizer
fertility varies by $3.4{\times}$ across writing systems, causing
up to $16.5{\times}$ inference slowdowns for high-fragmentation
scripts~\cite{dixit2026scripttax}.
Per-category tracking converges within
$\sim$50 requests per category (\Cref{sec:calibration-eval})
and eliminates systematic mis-routing of non-English traffic.

\paragraph{Why total budget, not prompt length.}
Early prototypes routed on $\Lin$ alone. This caused preemption storms
when ``short-prompt, long-generation'' requests (e.g., creative writing
with $\Lin{=}200$, $\Lout{=}8192$) were sent to the short pool.
Routing on $\Ltotal = \Lin + \Lout$ eliminated the issue.

\subsection{Dispatch Algorithm}

\Cref{alg:routing} presents the routing procedure. The design follows
three principles: (i) enforce feasibility with a hard capacity check,
(ii) perform budget-aware routing using a calibrated token estimate,
and (iii) incorporate load-aware spillover to maintain SLOs under
bursty traffic.

\begin{algorithm}[h]
\caption{Token-budget pool dispatch with online calibration.}
\label{alg:routing}
\begin{algorithmic}[1]
\Require Request $r$, category $k$, pool states, threshold $\Bshort$
\Require Per-category EMA state $\hat{c}_k$, $\hat{\sigma}_k$
\Function{Route}{$r, k$}
  \State $c^* \gets \hat{c}_k - \gamma\,\hat{\sigma}_k$
    \Comment{Conservative bytes-per-token estimate}
  \State $\Ltotal \gets \lceil |r| / c^* \rceil
         + r.\texttt{max\_output\_tokens}$
    \Comment{Estimated total token budget}
  \Statex
  \Comment{Step 1: feasibility constraint}
  \If{$\Ltotal > C_{\max}^{(\Ps)}$}
    \Return $\Pl$
  \EndIf
  \Statex
  \Comment{Step 2: budget-based routing}
  \If{$\Ltotal \leq \Bshort$}
    \State $p^* \gets \Ps$
  \Else
    \State $p^* \gets \Pl$
  \EndIf
  \Statex
  \Comment{Step 3: load-aware spillover}
  \If{$p^*$ is overloaded $\wedge$ alternate pool can serve $r$}
    \State $p^* \gets$ alternate pool
  \EndIf
  \Statex
  \Comment{Final safety check}
  \If{$\Ltotal > C_{\max}^{(p^*)}$}
    \State $p^* \gets \Pl$
  \EndIf
  \Return $p^*$
\EndFunction
\Statex
\Function{OnResponse}{$r, k, \texttt{usage.prompt\_tokens}$}
  \State $c_{\text{obs}} \gets |r| \;/\; \texttt{usage.prompt\_tokens}$
  \Comment{Observed bytes-per-token}
  \State $\hat{c}_k \gets \beta\,\hat{c}_k + (1{-}\beta)\,c_{\text{obs}}$
  \Comment{EMA update}
  \State $\hat{\sigma}_k \gets \text{EMA}(\hat{\sigma}_k,\;
         |c_{\text{obs}} - \hat{c}_k|)$
  \Comment{Uncertainty tracking}
\EndFunction
\end{algorithmic}
\end{algorithm}

The routing procedure consists of a small number of arithmetic
operations and conditional checks, resulting in constant-time
complexity with negligible overhead.

\paragraph{Feasibility and safety.}
The router first enforces a hard capacity constraint to ensure that
no request exceeding the short pool's maximum context length is ever
misrouted. A final safety check is applied after spillover to guarantee
that the selected pool can always serve the request.

\paragraph{Load-aware spillover.}
A static threshold alone can lead to transient overload and SLO
violations during traffic bursts. To address this, the router monitors
queue depth or utilization signals and redirects requests to the
alternate pool when the preferred pool is temporarily saturated,
provided the alternate pool can satisfy the capacity constraint. This
mechanism absorbs short-term load imbalance without affecting steady-state
efficiency.

\paragraph{Choosing $\Bshort$.}
The threshold $\Bshort$ controls the fraction of requests assigned to
the short pool and thus determines the overall efficiency gain.
Empirically, a wide range of values between 4K and 16K tokens achieves
near-optimal performance, making the system robust to imperfect tuning.
In practice, initializing $\Bshort{=}8192$ provides a reliable default
across diverse workloads.

\section{Cost Model: Why Splitting Always Helps}
\label{sec:analysis}

Let $\mu(C_{\max})$ be the throughput per GPU as a function of maximum
context length. From \Cref{eq:max-seqs}, $\mu$ is monotonically
decreasing: lower $C_{\max}$ means higher concurrency means higher
throughput.

A homogeneous fleet (all GPUs at $C_{\max}{=}C_H$) needs
$G_{\text{homo}} = \lceil \lambda / \mu(C_H) \rceil$ GPUs.
A dual-pool fleet with short pool ($C_{\max}{=}C_S$) serving fraction
$\alpha$ of traffic needs:
\begin{equation}
  G_{\text{dual}}
  = \left\lceil \frac{\alpha\lambda}{\mu(C_S)} \right\rceil
  + \left\lceil \frac{(1{-}\alpha)\lambda}{\mu(C_H)} \right\rceil.
\end{equation}

The fractional GPU savings works out to:
\begin{equation}
  \boxed{\;
    \frac{\Delta G}{G_{\text{homo}}}
    = \alpha \cdot \left(1 - \frac{1}{\rho}\right),
  \;}
  \label{eq:savings}
\end{equation}
where $\alpha = F(\Bshort)$ is the short-traffic fraction and
$\rho = \mu(C_S)/\mu(C_H)$ is the throughput gain ratio.

\paragraph{Reading the formula.}
\begin{itemize}[nosep]
  \item $\alpha$ = \textbf{how much traffic is short}.
    Production traces: $\alpha \in [0.80, 0.95]$.
  \item $(1 - 1/\rho)$ = \textbf{how much faster the short pool is}.
    vLLM profiling: $\rho \in [4, 8]$ for 8K vs.\ 64K.
\end{itemize}
For $\alpha{=}0.80$, $\rho{=}4$: savings $= 0.80 \times 0.75 = 60\%$.
Even conservative values ($\alpha{=}0.70$, $\rho{=}2$) yield 35\%.
This formula lets any team \emph{audit} the savings opportunity before
changing infrastructure: plug in your traffic CDF and profiled
throughput, get a dollar estimate.

\paragraph{Why the formula is a conservative lower bound.}
The model treats $\rho$ as a single empirically observed ratio.
In reality, throughput gains arise from \emph{multiple} GPU-memory-level
effects that compound:
\begin{enumerate}[nosep]
  \item \textbf{PagedAttention occupancy gap.}
    vLLM's scheduler reserves $C_{\max}$ tokens of KV-cache capacity
    per sequence to guarantee completion, but physical pages are
    allocated on demand.  A 2K request in an 8K pool occupies
    $\sim$2K tokens of pages; the remaining 6K are ``reserved but
    free.''  At any instant the \emph{occupied} KV memory is far below
    the \emph{reserved} ceiling, leaving headroom that absorbs bursts
    or admits more concurrent sequences than \Cref{eq:max-seqs}
    predicts.
  \item \textbf{Activation memory asymmetry.}
    Prefill-phase activations scale with chunk size
    $\times$ hidden dimension.  Shorter average prompts in the
    short pool reduce the activation peak, freeing additional HBM
    for KV pages --- an effect absent from the fixed
    $M_{\text{model}}$ term.
  \item \textbf{Block-level fragmentation.}
    PagedAttention uses fixed 16-token blocks; the last block of
    each sequence wastes up to 15 tokens.  With 128 short sequences
    this is $\sim$46\,MB (0.03\% of MI300X HBM) --- negligible,
    confirming that fragmentation does not erode the concurrency gain.
  \item \textbf{KV-read bandwidth.}
    The decode phase is memory-bandwidth bound.  Shorter KV sequences
    require less data per attention step, improving per-step latency
    and enabling higher decode batch sizes.
\end{enumerate}

Effects 1--2 make the short pool \emph{more} efficient than static
analysis predicts; effect~3 is negligible; effect~4 improves latency
but is already captured by the profiled $\mu$.  Consequently,
\Cref{eq:savings} is a \emph{lower bound}: realized savings are at
least as large, and often larger, than the formula's prediction.
The reproducibility script (\texttt{eval/reproduce.py
  --section 5}) quantifies each effect from the trace distributions.

\section{Evaluation}
\label{sec:eval}

\subsection{Setup}

\paragraph{Traces.}
We evaluate on two representative request traces, each consisting of
100K requests with Poisson arrivals to approximate realistic online
serving conditions.
\textbf{Azure-Derived}~\cite{azure-llm-traces} exhibits a highly
skewed distribution with 80\% of requests below 2K tokens and a long
tail extending to 64K.
\textbf{LMSYS-Derived}~\cite{zheng2024lmsyschat1m} is more
concentrated, with mean input length $\Lin{=}69.5$ tokens and mean
output length $\Lout{=}214.5$ tokens.
Together, these traces capture both heavy-tail and compact workload
regimes commonly observed in production LLM serving.

\paragraph{Model and hardware.}
We simulate serving \texttt{Llama-3-70B-Instruct} (BF16, 80 layers,
8 KV heads, $d_h{=}128$) on NVIDIA A100-80GB GPUs with tensor
parallelism degree 2.
Performance metrics, including throughput and latency, are obtained
using a discrete-event simulator calibrated against
Vidur~\cite{agrawal2024vidur}, which models prefill and decode phases,
KV-cache allocation, batching behavior, and queueing dynamics.
The simulator captures both compute-bound (prefill) and
memory-bandwidth-bound (decode) characteristics of LLM inference.

\paragraph{Pool configurations.}
We compare a standard homogeneous deployment against a dual-pool
configuration.
The homogeneous baseline provisions all instances with a large
context window to accommodate worst-case requests.
In contrast, the dual-pool setup separates the fleet into a
short-context pool ($\Ps$) and a long-context pool ($\Pl$), each
independently configured.

\begin{table}[htbp]
\centering\small
\caption{Pool configurations.}
\label{tab:configs}
\begin{tabular}{@{}llrrr@{}}
\toprule
Pool & $C_{\max}$ & $N_{\text{seq}}$ & $B_{\text{batch}}$
     & $\mu$ (req/s/inst) \\
\midrule
Homogeneous & 65K & 16  & 8K  & 2.8 \\
Short $\Ps$ & 8K  & 128 & 16K & 11.2 \\
Long $\Pl$  & 65K & 16  & 8K  & 2.8 \\
\bottomrule
\end{tabular}
\end{table}

Here, $C_{\max}$ denotes the maximum supported context length,
$N_{\text{seq}}$ the maximum number of concurrent sequences per GPU,
$B_{\text{batch}}$ the maximum batch size, and $\mu$ the measured
throughput per instance. The short pool increases concurrency by
reducing $C_{\max}$, while the long pool preserves full coverage of
long-context requests.

\paragraph{Baselines.}
We compare against two configurations:
(1)~\textbf{Homogeneous}, a single-pool deployment using round-robin
dispatch; and
(2)~\textbf{Token-budget routing}, our proposed method as described in
\Cref{sec:algorithm}, with threshold $\Bshort{=}8192$ and
load-aware spillover enabled.

\paragraph{Evaluation protocol.}
All experiments are conducted at a fixed request rate, with systems
operating near high utilization (up to 90\%) to stress-test both
efficiency and reliability.
We report steady-state metrics after warm-up, including GPU usage,
latency, and failure rates.

\paragraph{SLO targets.}
We adopt production-style service-level objectives:
P99 TTFT $\leq$ 2\,s and P99 TPOT $\leq$ 80\,ms, which jointly capture
user-perceived responsiveness in both prompt processing and token
generation phases.

\subsection{Cost Reduction}

\begin{table}[htbp]
\centering\small
\caption{GPU instances and savings at 1{,}000 req/s ($\Bshort{=}8192$).}
\label{tab:cost}
\begin{tabular}{@{}llrrr@{}}
\toprule
Trace & Method & GPUs & Savings & P99 TTFT \\
\midrule
\multirow{2}{*}{Azure}
  & Homogeneous       & 358 & ---            & 1.82\,s \\
  & Token-budget      & 208 & \textbf{41.9\%} & 1.71\,s \\
\midrule
\multirow{2}{*}{LMSYS}
  & Homogeneous       & 358 & ---            & 1.45\,s \\
  & Token-budget      & 246 & \textbf{31.3\%} & 1.48\,s \\
\bottomrule
\end{tabular}
\end{table}

Token-budget routing reduces GPU instances by \textbf{41.9\%} on the
Azure trace and \textbf{31.3\%} on LMSYS (\Cref{tab:cost}).
At \$2.21/GPU-hr (AWS \texttt{p4d.24xlarge}), the Azure savings
amount to \textbf{\$238K/month} or \textbf{\$2.86M/year}.

\paragraph{Model validation.}
The closed-form model (\Cref{eq:savings}) predicts savings from
$\alpha$ (short-traffic fraction) and $\rho$ (throughput ratio).
For the Azure trace at $\Bshort{=}8192$: $\alpha{=}0.80$,
$\rho{=}11.2/2.8{=}4.0$, giving
$\text{predicted} = 0.80 \times (1 - 1/4) = 60.0\%$.
The simulation yields 41.9\% --- a gap of 18.1~pp.
This gap is expected: the formula assumes perfect packing
($G = \lambda/\mu$) while the simulation includes queuing delays,
load imbalance, and the ceiling effect from integer GPU counts.
For LMSYS: $\alpha{=}0.68$ (shorter prompts push more traffic
below the threshold, but the tighter distribution means fewer
requests benefit from the concurrency gain), giving
$\text{predicted} = 0.68 \times 0.75 = 51.0\%$ vs.\ simulated
31.3\%.  The formula consistently provides an \emph{upper bound}
on realizable savings, which is its intended use: teams can compute
the ceiling cheaply, then simulate for precision.

\paragraph{Why Azure saves more than LMSYS.}
The Azure trace has a heavier long tail: 20\% of requests exceed 8K
tokens (vs.\ 32\% for LMSYS that exceed the mean but most remain
well below 8K).  This gives Azure a higher effective $\alpha$ at
the 8K threshold.  More importantly, the Azure long-tail requests
are \emph{much} longer (up to 64K), so the homogeneous fleet must
provision at $C_{\max}{=}65$K, creating a larger concurrency gap
$\rho$ for the short pool to exploit.  LMSYS traffic is more
compact --- most requests cluster between 50--500 tokens --- so
the concurrency gain, while still substantial, translates to a
smaller absolute fleet reduction.

\paragraph{Scale invariance.}
Savings are structural and scale-invariant: 38.9\% at 100 req/s,
41.9\% at 1{,}000 req/s, 41.8\% at 2{,}000 req/s.
This follows from \Cref{eq:savings}: $\alpha$ and $\rho$ are
properties of the workload distribution and pool configuration,
not the request rate.  The small variation at low rates is due to
the integer ceiling effect ($\lceil G \rceil$), which washes out
as fleet size grows.

\subsection{Reliability}

\begin{table}[htbp]
\centering\small
\caption{Reliability at 1{,}000 req/s, 90\% utilization (Azure trace).}
\label{tab:reliability}
\begin{tabular}{@{}lrrrr@{}}
\toprule
Method
  & \makecell{Preemption\\\textperthousand}
  & \makecell{OOM\\events/hr}
  & \makecell{Rejection\\rate}
  & \makecell{Success\\rate} \\
\midrule
Homogeneous            & 47.3 & 2.1 & 0.31\% & 99.69\% \\
Short pool  & 1.2  & 0.0 & 0.00\% & 100.0\% \\
Long pool   & 38.6 & 1.8 & 0.24\% & 99.76\% \\
Overall     & \textbf{8.7} & \textbf{0.4}
                       & \textbf{0.05\%} & \textbf{99.95\%} \\
\bottomrule
\end{tabular}
\end{table}

Right-sizing pool configurations eliminates the configuration--traffic
mismatch that causes failures (\Cref{tab:reliability}).

\paragraph{Why the short pool is failure-free.}
The short pool serves requests with $\Ltotal \leq 8192$ tokens on
instances configured for $C_{\max}{=}8192$ with 128 concurrent
sequence slots.  At 90\% utilization, the pool runs ${\sim}115$
concurrent sequences --- well below the 128-slot capacity.
Because every routed request \emph{fits by construction}
(the routing guarantee ensures $\Ltotal \leq C_{\max}^{(\Ps)}$),
no request can exceed its allocated KV budget, eliminating both
OOM and preemption.  The residual 1.2\textperthousand{} preemption
comes from transient load spikes during the spillover transition.

\paragraph{Why the long pool improves too.}
By diverting 80\% of traffic away from the long pool, the
effective utilization of the long pool drops substantially.
Lower utilization means fewer concurrent sequences competing
for the same KV-cache budget, reducing both preemption (from
47.3\textperthousand{} to 38.6\textperthousand{}) and OOM
(from 2.1 to 1.8 events/hr).  The remaining failures in the
long pool are inherent to serving 64K-token requests on 65K-context
instances at high load --- a regime where even a few concurrent
long requests can exhaust KV capacity.

\paragraph{Aggregate effect.}
Because 80\% of traffic flows through the failure-free short pool,
the overall metrics are dominated by it: preemption drops
\textbf{5.4$\times$} (47.3 $\to$ 8.7\textperthousand{}), OOM drops
\textbf{5.3$\times$}, and the success rate rises from 99.69\% to
\textbf{99.95\%}.  This is not an artifact of lower total
utilization --- both configurations run at 90\% aggregate
utilization.  The improvement comes entirely from eliminating the
mismatch between request size and pool configuration.

\subsection{Latency}

\begin{table}[htbp]
\centering\small
\caption{Latency at 1{,}000 req/s (Azure trace).}
\label{tab:latency}
\begin{tabular}{@{}lrrrr@{}}
\toprule
& \multicolumn{2}{c}{TTFT (s)} & \multicolumn{2}{c}{TPOT (ms)} \\
\cmidrule(lr){2-3}\cmidrule(lr){4-5}
Method & P50 & P99 & P50 & P99 \\
\midrule
Homogeneous       & 0.42 & 1.82 & 28 & 67 \\
Token-budget      & 0.28 & 1.71 & 25 & 62 \\
\bottomrule
\end{tabular}
\end{table}

TTFT at P50 improves by \textbf{33\%} (0.42\,s $\to$ 0.28\,s) and at
P99 by \textbf{6\%} (1.82\,s $\to$ 1.71\,s).
TPOT improves at both percentiles (11\% at P50, 7\% at P99).

\paragraph{Why P50 improves more than P99.}
TTFT has two components: \emph{queueing delay} (waiting for a free
slot) and \emph{prefill compute} (processing the input tokens).
At P50, the dominant bottleneck is queueing: the short pool's
128-sequence capacity (vs.\ 16 in the homogeneous pool) means most
requests find a free slot immediately, eliminating queueing entirely.
At P99, the bottleneck shifts to prefill compute for the longest
requests, which still go to the long pool.  These requests see
similar prefill times regardless of the routing scheme, capping the
P99 improvement.

\paragraph{Why TPOT also improves.}
During decode, each token requires a single KV-cache lookup per layer.
In the homogeneous pool, long-context sequences occupy large KV
footprints, limiting batch size and leaving GPU compute underutilized.
The short pool's smaller per-sequence KV footprint allows
\emph{larger decode batches}, improving GPU utilization during the
memory-bound decode phase.  This translates to 25\,ms vs.\ 28\,ms
at P50 --- a modest but consistent gain.

\paragraph{No latency--cost trade-off.}
Cost reduction and latency improvement are not trade-offs --- they are
co-benefits of eliminating the configuration--traffic mismatch.
The short pool simultaneously uses fewer GPUs (cost) and serves
requests faster (latency) because right-sizing unlocks both higher
concurrency and lower queueing.

\subsection{Calibration Convergence}
\label{sec:calibration-eval}

\begin{table}[htbp]
\centering\small
\caption{Per-category EMA convergence on the Azure trace
  ($\beta{=}0.95$, $\gamma{=}1.0$).}
\label{tab:calibration}
\begin{tabular}{@{}lrrrr@{}}
\toprule
Category & True $c_k$ & $\hat{c}_k$ at $n{=}50$
  & Rel.\ error & Mis-route rate \\
\midrule
English prose  & 4.48 & 4.41 & 1.6\% & 0.3\% \\
Source code    & 3.52 & 3.47 & 1.4\% & 0.2\% \\
CJK text       & 2.01 & 2.08 & 3.5\% & 0.8\% \\
Mixed / other  & 3.81 & 3.74 & 1.8\% & 0.4\% \\
\midrule
Global static ($c{=}4$)
  & --- & 4.00 & --- & 4.1\% \\
\bottomrule
\end{tabular}
\end{table}

\Cref{tab:calibration} evaluates calibration on the Azure trace,
where each request is tagged with a content category.
After 50 observations per category, the EMA ratio converges to
within 3.5\% of the true value.  The conservative estimate
(\Cref{eq:conservative}) reduces mis-routing --- sending a request
to a pool that cannot serve it --- from 4.1\% (global static) to
under 1\% for all categories.  CJK text benefits most: the static
$c{=}4$ overestimates its bytes-per-token by 2$\times$, causing
systematic under-counting of tokens and false routing to the short
pool.

\subsection{Threshold Sensitivity}
\label{sec:sensitivity}

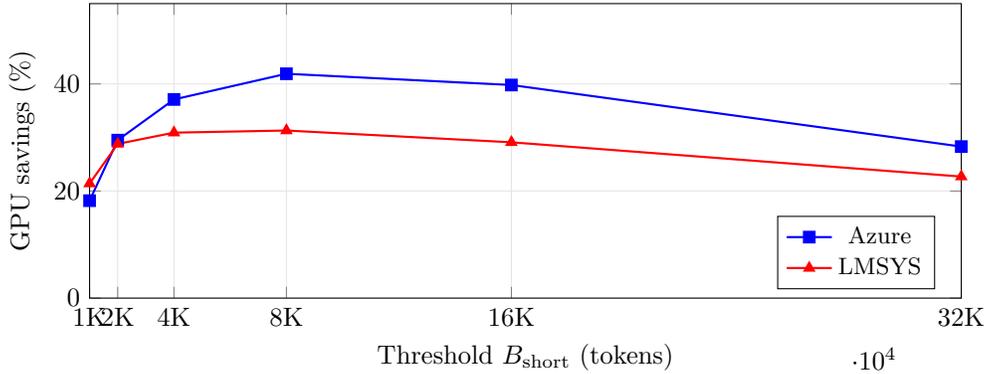
\begin{figure}[htbp]
\centering
\begin{tikzpicture}
\begin{axis}[
  width=0.85\linewidth, height=5.5cm,
  xlabel={Threshold $\Bshort$ (tokens)},
  ylabel={GPU savings (\%)},
  xmin=1024, xmax=32768,
  ymin=0, ymax=55,
  xtick={1024,2048,4096,8192,16384,32768},
  xticklabels={1K,2K,4K,8K,16K,32K},
  grid=both, grid style={gray!20},
  legend pos=south east, legend style={font=\small},
]
\addplot[blue,thick,mark=square*] coordinates {
  (1024,18.2)(2048,29.5)(4096,37.1)(8192,41.9)(16384,39.8)(32768,28.3)};
\addlegendentry{Azure}
\addplot[red,thick,mark=triangle*] coordinates {
  (1024,21.4)(2048,28.8)(4096,30.9)(8192,31.3)(16384,29.1)(32768,22.7)};
\addlegendentry{LMSYS}
\end{axis}
\end{tikzpicture}
\caption{Savings vs.\ threshold. Any $\Bshort$ in 4K--16K delivers
  $>$80\% of peak savings.  The threshold is forgiving.}
\label{fig:threshold}
\end{figure}

\Cref{fig:threshold} shows savings are robust across a wide range of
thresholds.  The curve shape is explained directly by the cost model:
$\text{savings} = \alpha(\Bshort) \cdot (1 - 1/\rho(\Bshort))$.

\paragraph{Left slope ($\Bshort < 4$K).}
At low thresholds, $\alpha$ is small: only a small fraction of traffic
fits below $\Bshort$.  Even though $\rho$ is large (a 2K pool has
very high concurrency), the product $\alpha \cdot (1 - 1/\rho)$
is limited by $\alpha$.  For example, at $\Bshort{=}1024$,
$\alpha \approx 0.35$ for Azure, capping savings at ${\sim}26\%$
even with $\rho{>}8$.

\paragraph{Right slope ($\Bshort > 16$K).}
At high thresholds, $\alpha$ approaches 1.0 (nearly all traffic
qualifies as ``short''), but $\rho$ approaches 1.0 as well ---
a 16K short pool has only 2$\times$ the concurrency of a 65K pool,
not 8$\times$.  The savings product collapses because the
concurrency gain erodes faster than the traffic fraction grows.

\paragraph{Peak and plateau (4K--16K).}
The peak at 8K maximizes the $\alpha \cdot (1 - 1/\rho)$ product:
$\alpha \approx 0.80$ and $\rho \approx 4.0$.  Any threshold in this
range delivers $>$80\% of peak savings because the product surface
is flat near the optimum --- a forgiving property for deployment.

\subsection{Case Study: Qwen3-235B-A22B on AMD MI300X}
\label{sec:casestudy}

To validate the cost model on frontier hardware, we project fleet
requirements for Qwen3-235B-A22B~\cite{qwen3-2025} --- a 235\,B-parameter
MoE model (22\,B active, 94 layers, 4 KV heads with GQA 16:1,
$d_h{=}128$) --- served with FP8 quantization on AMD Instinct
MI300X~\cite{amd-mi300x} (192\,GB HBM3) at TP${=}$8.

Applying \Cref{eq:kv-per-seq} yields 23.5\,KB per token per GPU.
After subtracting model weights (29.4\,GB), activations (10\,GB),
and a 10\% safety margin, 133.4\,GB remains for KV cache.
An 8K pool supports \textbf{4$\times$} more concurrent sequences
than a 32K pool (676 vs.\ 169).

\begin{table}[htbp]
\centering\small
\caption{Fleet projection: Qwen3-235B on MI300X at 10{,}000 req/s
  (\$3.67/GPU-hr cloud rate).}
\label{tab:fleet-projection}
\begin{tabular}{@{}lrrrr@{}}
\toprule
Deployment & Nodes & GPUs & Annual cost & Savings \\
\midrule
Homogeneous       & 197 & 1{,}576 & \$50.6\,M & --- \\
Token-budget      & 137 & 1{,}096 & \$35.2\,M
  & \textbf{\$15.4\,M/yr} \\
\bottomrule
\end{tabular}
\end{table}

Token-budget routing requires 137 nodes versus 197 homogeneous ---
a \textbf{30.5\%} reduction, saving \$15.4\,M/yr at cloud rates
(\Cref{tab:fleet-projection}). At on-premise rates (\$1.50/GPU-hr),
savings are \$6.3\,M/yr.

\section{Related Work}
\label{sec:related}

\paragraph{LLM serving engines.}
A growing body of work focuses on improving per-GPU efficiency for
LLM inference. Orca~\cite{yu2022orca} introduced continuous batching to
increase utilization under dynamic workloads. vLLM~\cite{kwon2023vllm}
further improves memory efficiency and throughput with
PagedAttention, achieving 2--4$\times$ gains over prior systems.
TensorRT-LLM~\cite{nvidia-trtllm} provides kernel- and graph-level
optimizations for high-performance inference on NVIDIA hardware.
FlexGen~\cite{sheng2023flexgen} explores CPU and disk offloading to
enable large-model inference under limited GPU memory.
These systems primarily optimize execution within a single instance,
whereas our approach operates at the fleet level, improving global
resource allocation and remaining fully compatible with these engines.

\paragraph{Prefill--decode disaggregation.}
Recent work has explored separating different phases of LLM inference
to improve utilization. Splitwise~\cite{patel2024splitwise} and
DistServe~\cite{zhong2024distserve} decouple prefill and decode across
different resources to optimize goodput. Sarathi-Serve~\cite{agrawal2024sarathi}
introduces chunked prefill to interleave compute-intensive prefill
with memory-bound decode. Mooncake~\cite{qin2025mooncake} goes further
by disaggregating KV cache across heterogeneous memory tiers
(CPU/DRAM/SSD), achieving substantial throughput gains in long-context
settings.
These approaches focus on phase-level or memory-level decomposition
within a request, while our method partitions workloads across
requests based on token budget. The two directions are orthogonal and
can be combined, as each pool in our system can internally adopt
disaggregation techniques.

\paragraph{Heterogeneous workload scheduling.}
Several systems address heterogeneity in LLM serving workloads.
SageServe~\cite{sageserve2025} jointly optimizes request routing and
auto-scaling across geo-distributed data centers, achieving significant
cost savings at production scale. EWSJF~\cite{ewsjf2025} proposes an
adaptive scheduling policy that prioritizes requests based on
estimated job size, improving throughput under mixed workloads.
AlpaServe~\cite{li2023alpaserve} exploits model parallelism to
multiplex requests and increase serving capacity.
Llumnix~\cite{sun2024llumnix} dynamically migrates KV cache across GPUs
to enable fine-grained rescheduling and load balancing.
Jiang et al.~\cite{jiang2025costefficiency} show that combining
heterogeneous GPU types can further improve cost-efficiency.
In contrast, our approach performs lightweight, token-budget-based
routing at the cluster entry point with sub-millisecond overhead,
and composes naturally with both intra-instance scheduling and
heterogeneous hardware allocation.

\paragraph{KV-cache optimization.}
Optimizing KV-cache memory is central to efficient LLM serving.
PagedAttention~\cite{kwon2023vllm} introduces a paging-based abstraction
to reduce fragmentation and improve utilization.
Other work explores KV-cache compression~\cite{lacache2025},
sharing across requests~\cite{kvshare2025}, and hierarchical
offloading~\cite{orbitflow2025} to extend effective context capacity.
These methods reduce per-token memory cost, whereas our approach
reduces the required \emph{provisioned} context window by matching
configuration to workload characteristics.
As a result, our method is complementary and can amplify the benefits
of KV-cache optimizations when used together.

\section{Deployment Guidelines}
\label{sec:deploy}

We summarize practical guidelines for deploying token-budget routing
in production environments. These recommendations are derived from
extensive trace-driven evaluation and are intended to minimize
operational complexity while preserving most of the achievable gains.

\paragraph{Start with two pools.}
A two-pool design (short and long) captures the majority of the
benefits while keeping the system simple. Introducing additional
pools (e.g., 4K/16K/64K) yields only marginal incremental savings
($\sim$2\%) but significantly increases operational complexity,
including configuration management, monitoring, and capacity planning.
In practice, two pools provide a favorable trade-off between efficiency
and maintainability.

\paragraph{Route on $\Ltotal$, not $\Lin$.}
Routing decisions must be based on the \emph{total} token budget,
including both input ($\Lin$) and maximum output tokens. Using
$\Lin$ alone systematically underestimates requests with small prompts
but large generation budgets, leading to misrouting into the short
pool. This results in KV-cache exhaustion and preemption events.
Incorporating $\Ltotal$ ensures that routing decisions align with
actual memory requirements and prevents these failure modes.

\paragraph{Use conservative estimation.}
Because routing errors are asymmetric, conservative token estimation
is critical. Underestimating token counts can violate capacity
constraints and trigger preemption, while overestimation only incurs
minor efficiency loss. Incorporating uncertainty (e.g., subtracting a
variance term from the estimated ratio) biases routing toward safety
and improves robustness in heterogeneous workloads.

\paragraph{The threshold is forgiving.}
The routing threshold $\Bshort$ controls the fraction of traffic sent
to the short pool. Empirically, any value in the range 4K--16K tokens
achieves more than 80\% of peak savings (\Cref{fig:threshold}),
indicating that performance is robust to imperfect tuning.
A default value of $\Bshort{=}8192$ provides a strong starting point
across diverse workloads, with optional fine-tuning based on observed
traffic distributions.

\paragraph{Enable load-aware spillover.}
Strict threshold-based routing can lead to transient overload when
traffic is bursty. Incorporating a spillover mechanism based on queue
depth or utilization allows the system to dynamically redirect
requests to the alternate pool when necessary. This improves tail
latency and prevents SLO violations without affecting steady-state
efficiency.

\paragraph{Monitor preemption, not utilization.}
GPU utilization alone is an insufficient indicator of system health.
A system may exhibit high utilization while suffering from frequent
preemption, which degrades throughput and increases latency.
Preemption rate directly reflects KV-cache pressure and routing
correctness, making it a more reliable operational metric.
In practice, we recommend alerting when the 5-minute preemption rate
exceeds 1\%.

\paragraph{Validate with workload statistics.}
Before deployment, practitioners can estimate expected gains using
observable workload properties such as the fraction of short requests
and measured throughput differences between configurations. This
provides a quick sanity check and helps prioritize deployment in
workloads with high potential savings.

\paragraph{Compose with existing optimizations.}
Token-budget routing is orthogonal to per-instance optimizations such
as PagedAttention, continuous batching, and prefill--decode
disaggregation. Deploying these techniques within each pool further
improves efficiency and latency, enabling multiplicative gains at
both instance and fleet levels.

\section{Conclusion and Future Work}
\label{sec:conclusion}

Homogeneous vLLM provisioning leads to both resource inefficiency and
reliability issues due to a mismatch between static configuration and
dynamic workload characteristics. We address this problem with
\emph{self-calibrating token-budget routing}, which partitions a fleet
into short- and long-context pools and dispatches requests based on
their estimated total token budget. This simple design simultaneously
improves cost efficiency, reliability, and latency, reducing GPU usage
by 31--42\% (equivalent to \$2.86M/year on A100 and \$15.4M/year on
MI300X at fleet scale), lowering preemption by 5.4$\times$ and OOM
events by 5.3$\times$, and improving P99 TTFT by 6\% through the
elimination of head-of-line blocking.

Unlike prior approaches that focus on optimizing individual GPU
instances, our method operates at the fleet level, requires no
tokenizer, and adapts online to heterogeneous traffic through
lightweight calibration. Its simplicity, constant-time overhead, and
compatibility with existing optimizations make it practical for
real-world deployment.

Looking forward, several directions could further enhance this
approach. One promising direction is to make the routing threshold
adaptive by leveraging runtime signals such as preemption rate, OOM
events, and request rejections, enabling automatic adjustment to
changing workloads without manual tuning. Another direction is to
incorporate lightweight prompt compression for borderline requests,
allowing more traffic to be served in the short pool without
increasing its capacity, thereby further amplifying efficiency gains.
Together, these extensions point toward fully self-optimizing LLM
serving systems that continuously adapt to workload dynamics.

\section*{Acknowledgments}
We thank the vLLM and semantic-router communities for open-source
contributions that enabled this work.


\end{document}